\documentclass[a4paper]{article}

\usepackage[margin=1in]{geometry} % full-width

\usepackage{setspace}
\usepackage{float} 

\usepackage{amsmath}
\usepackage{amsthm}
\usepackage{amssymb}
\usepackage{bbm}

\usepackage{subcaption}
\usepackage{multirow}
\usepackage{newfloat}
\usepackage{wrapfig}
\usepackage{caption} % for customizing caption format
\usepackage{hyperref}

\usepackage[table,xcdraw]{xcolor}

\usepackage[utf8]{inputenc}

\usepackage[sort&compress]{natbib}
\theoremstyle{plain}

\theoremstyle{definition}

\usepackage{graphicx, color}
\graphicspath{{fig/}}

\usepackage{amsmath, amssymb, amsfonts, mathtools}
\usepackage{physics}
\usepackage{cancel}
\usepackage{thmtools, thm-restate}
\usepackage{accents}

\usepackage{pict2e, picture}
\makeatletter
\DeclareRobustCommand{\Arrow}[1][]{%
\check@mathfonts
\if\relax\detokenize{#1}\relax
\settowidth{\dimen@}{$\m@th\rightarrow$}%
\else
\setlength{\dimen@}{#1}%
\fi
\sbox\z@{\usefont{U}{lasy}{m}{n}\symbol{41}}%
\begin{picture}(\dimen@,\ht\z@)
\roundcap
\put(\dimexpr\dimen@-.7\wd\z@,0){\usebox\z@}
\put(0,\fontdimen22\textfont2){\line(1,0){\dimen@}}
\end{picture}%
}
\makeatother

\newcommand{\zA}{\boldsymbol{A}}
\newcommand{\zB}{\boldsymbol{B}}
\newcommand{\zC}{\boldsymbol{C}}
\newcommand{\zD}{\boldsymbol{D}}

\newcommand{\zF}{\boldsymbol{F}}

\newcommand{\zu}{\boldsymbol{u}}

\newcommand{\R}{\mathbb{R}}

\DeclareMathAlphabet{\nummathbb}{U}{BOONDOX-ds}{m}{n}

\makeatletter
\DeclareRobustCommand\widecheck[1]{{\mathpalette\@widecheck{#1}}}
\def\@widecheck#1#2{%
    \setbox\z@\hbox{\m@th$#1#2$}%
    \setbox\tw@\hbox{\m@th$#1%
       \widehat{%
          \vrule\@width\z@\@height\ht\z@
          \vrule\@height\z@\@width\wd\z@}$}%
    \dp\tw@-\ht\z@
    \@tempdima\ht\z@ \advance\@tempdima2\ht\tw@ \divide\@tempdima\thr@@
    \setbox\tw@\hbox{%
       \raise\@tempdima\hbox{\scalebox{1}[-1]{\lower\@tempdima\box
\tw@}}}%
    {\ooalign{\box\tw@ \cr \box\z@}}}
\makeatother

\title{Towards trustworthy seizure onset detection using workflow notes}

\author{Khaled Saab,$^{1}$ Siyi Tang,$^{1}$ Mohamed Taha,$^{2}$ \\
Christopher Lee-Messer,$^{3,*}$ Christopher R\'e,$^{4,*}$ and Daniel L. Rubin$^{5,*}$}

\date{
	$^1$Department of Electrical Engineering, Stanford University, Stanford, CA, USA \\
        $^2$Department of Neurology, Stanford University, Stanford, CA, USA \\
        $^3$Department of Child Neurology, Stanford University, Stanford, CA, USA \\
	$^4$Department of Computer Science, Stanford University, Stanford, CA, USA \\
	$^5$Department of Biomedical Data Science, Radiology, and Medicine, Stanford University, Stanford, CA, USA
}

\begin{document}
\maketitle
\pagenumbering{gobble}

$^{*}$Equal contribution. 

Address correspondence to Khaled Saab (ksaab@stanford.edu)
% \newline

% \indent Khaled Saab \\
% \indent ksaab@stanford.edu \\
% \indent Department of Electrical Engineering \\
% \indent Stanford University, CA, USA
% \newline

% \indent Daniel L. Rubin \\
% \indent dlrubin@stanford.edu \\
% \indent Department of Biomedical Data Science, Radiology, and Medicine \\
% \indent Stanford University, CA, USA

% \newpage
\pagenumbering{arabic}
\setcounter{page}{1}

\section*{Abstract}
A major barrier to deploying healthcare AI models is their trustworthiness. One form of trustworthiness is a model's robustness across different subgroups: while existing models may exhibit expert-level performance on aggregate metrics, they often rely on non-causal features, leading to errors in hidden subgroups.
To take a step closer towards trustworthy seizure onset detection from EEG, we propose to leverage annotations that are produced by healthcare personnel in routine clinical workflows -- which we refer to as workflow notes -- that include multiple event descriptions beyond seizures. 
Using workflow notes, we first show that by scaling training data to an unprecedented level of $68{,}920$ EEG hours, seizure onset detection performance significantly improves (+$12.3$ AUROC points) compared to relying on smaller training sets with expensive manual gold-standard labels. 
Second, we reveal that our binary seizure onset detection model underperforms on clinically relevant subgroups (e.g., up to a margin of $6.5$ AUROC points between pediatrics and adults), while having significantly higher false positives on EEG clips showing non-epileptiform abnormalities compared to any EEG clip (+$19$ FPR points).
To improve model robustness to hidden subgroups, we train a multilabel model that classifies $26$ attributes other than seizures, such as spikes, slowing, and movement artifacts.
We find that our multilabel model significantly improves overall seizure onset detection performance (+$5.9$ AUROC points) while greatly improving performance among subgroups (up to +$8.3$ AUROC points), and decreases false positives on non-epileptiform abnormalities by $8$ FPR points.
Finally, we propose a clinical utility metric based on false positives per $24$ EEG hours and find that our multilabel model improves this clinical utility metric by a factor of $2\times$ across different seizure onset detection recall and latency times. 
These results demonstrate the importance of leveraging additional cost-effective supervision to improve model robustness to classification errors in patient subgroups.
\newpage
\section{Introduction}

The scalp electroencephalogram (EEG) is a non-invasive and valuable technique to measure the brain’s electrical activity. Unlike other modalities that image the brain (e.g., fMRI, PET), EEG enables continuous analysis of rapid changes in the brain’s electrical activity. In the intensive care unit (ICU), EEG is critical for the detection of seizures that may lack a behavioral correlate and worsen brain injury. Moreover, EEG is an essential tool to diagnose and care for epileptic patients of all ages \citep{schomer2012niedermeyer}. 

While analyzing EEG data is a critical healthcare task, it poses several challenges. First, the continuous recording of hours of multi-channel EEG results in a vast amount of data that requires thorough interpretation, which is a highly time-consuming and costly task that demands deep neurologic-epileptologic understanding. 
Second, the gold-standard for EEG analysis is done by fellowship trained clinical neurophysiologists, who have not only been trained to identify seizure patterns, but also many common artifacts. 
For example, common artifacts on EEG signals may include muscle movement or environment noise, along with countless non-epileptiform abnormalities such as spikes and slowing.
Finally, there is a shortage of EEG specialists, and as a result, low resource communities lack access to EEG interpretation \cite{brogger2018visual}. Thus, there is a strong need to develop reliable tools that help clinicians analyze EEG data more efficiently.

Many studies have shown that deep learning (DL) techniques present great promise for automated seizure detection. 
There have been substantial efforts for curating large and publicly available EEG datasets, such as the Temple University Hospital Seizure Detection (TUSZ) corpus that includes thousands of EEGs from hundreds of patients \citep{obeid2016temple, shah2018temple}. The availability of large public datasets has enabled rapid progress in benchmarking and improved seizure detection models \citep{saab2020weak,tang2022spatiotemporal,li2020epileptic,thuwajit2021eegwavenet,ahmedt2020neural,golmohammadi2019automatic}. Recently, a DL model named SParCNet was trained on 6,097 EEGs from 2,711 patients, annotated independently by 20 fellowship-trained neurophysiologists, and was found to match or exceed most experts in classifying seizures \citep{jing2023development}.

Due to the high-stakes nature of healthcare, trustworthiness of DL models remains a major roadblock to clinical adoption \citep{wiens2019no,reus2023automated}. Alarmingly, there has been a growing body of work revealing that healthcare models with ``expert-level'' performance often rely on non-generalizable features \citep{degrave2021ai,badgeley2019deep}, resulting in unexpected drops in performance over hidden subgroups \citep{oakden2020hidden,saab2022reducing} or under data distribution shifts \citep{zech2018variable}. While many studies report impressive overall seizure detection performance \citep{tang2022spatiotemporal,jing2023development}, such studies lack the in-depth analysis needed to understand the clinically meaningful failure modes of existing models. For example, pediatric EEGs look drastically different from adult EEGs, different seizure types display unique EEG patterns, and there may be different types of abnormalities present in EEGs recorded from the ICU as compared to other clinical settings \citep{schomer2012niedermeyer}; as a result, models may underperform on specific age groups, seizure subtypes, or ICU patients. Unfortunately, conducting an in-depth error analysis requires manual interpretation of both EEGs and model predictions over a diverse set of studies, making it a costly process. However, a clear understanding of a model's systematic errors is critical to provide trust in model predictions for clinical adoption.

In this work, we provide a strategy to scale training data, conduct a subgroup robustness analysis, and improve the trustworthiness of seizure onset detection models in a cost-effective manner. 
As opposed to relying on expensive gold-standard labels, which require a fellow-trained neurophysiologist to label EEGs outside existing clinical workflows, we propose to leverage seizure annotations that are produced by healthcare personnel within existing clinical workflows \citep{saab2020weak} -- which we refer to as workflow notes.
Since workflow notes are produced as part of routine clinical practice, we are able to train our DL models on an unprecedented scale of $68,920$ EEG hours. 
To conduct an in-depth error analysis we stratify the evaluation set of EEG recordings into clinically-relevant subgroups and analyze discrepancies in seizure onset detection performance in each subgroup. 
In particular, we use a combination of patient metadata (e.g., age), expert-provided subgroup labels (e.g., seizure types), along with numerous EEG attributes, such as spikes, slowing, movements, jerks, photoelectric stimulation, hyperventilation, and more (full list in Supplementary Table \ref{table:regexs}), that are readily available from workflow notes. 

To improve model robustness to non-epileptiform abnormalities and hidden subgroups, we utilize the workflow notes to increase class specificity. 
Specifically, as opposed to training a binary classification model (seizure or no seizure onset), we train a multilabel model to classify $25$ classes in addition to seizure onset, such as spikes, slowing, and hyperventilation. 
Additionally, we study how our improvements in seizure onset detection robustness translate to clinical utility by tracking the false positives per $24$ hours for different deployment settings. 
\section{Results}

\begin{figure}[t]
  \centering
  \includegraphics[width=0.95\textwidth]{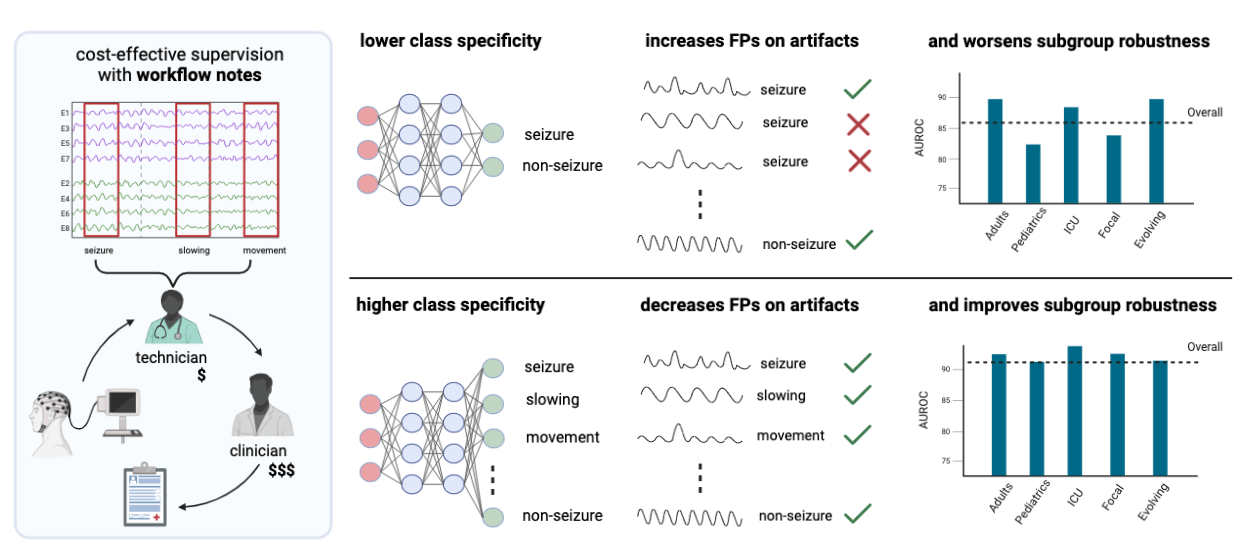}
  \caption{\textbf{Results overview:} We find that increasing class specificity by providing additional supervision decreases false positives on artifacts and improves subgroup robustness. Importantly, we supervise our models on large scale data ($68,920$ EEG hours) using readily available notes produced within clinical workflows (left panel).}
  \label{fig:figure1}
\end{figure}

We first describe how we utilize workflow notes to scale supervision to $68{,}920$ EEG hours ($4{,}125{,}225$ 60-sec EEG clips) in a cost-effective manner, and show that training a model to detect seizure onset using workflow notes greatly improves performance compared with a model trained with a smaller set of gold-standard, expert-labeled EEG clips (Section \ref{subsec:scaling_training}). We further utilize the workflow notes to reveal that even with large-scale training, our binary seizure onset detection model underperforms on clinically-relevant subgroups of patients, and has higher false positive rates for non-seizure EEG clips with abnormal patterns (Section \ref{subsec:revealing_subgroups}). To improve our model's performance across subgroups, we train a multilabel model to classify $25$ attributes extracted from the workflow notes, in addition to seizure onset (Section \ref{subsec:improving_subgroups},  Figure \ref{fig:figure1}). Finally, we propose a metric of clinical utility to assess the degree to which the multilabel model improves clinical utility over a range of settings (Section \ref{subsec:clinical}).

\begin{figure}[t]
  \centering
  \includegraphics[scale=0.25]{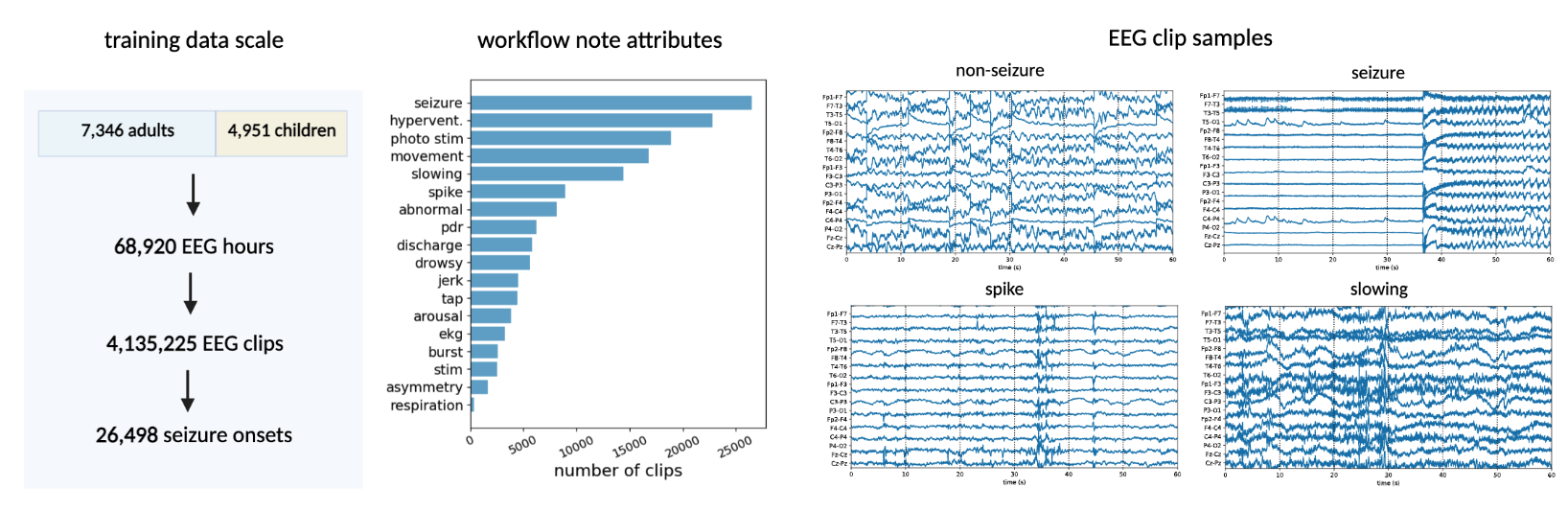}
  \caption{\textbf{Training dataset overview:} In the left panel, we provide statistics on the scale our training dataset of EEG recordings aggregated from adult and pediatric hospitals. In the middle panel, we plot the histogram of atribute labels extracted from workflow notes. In the right panel, we visualize four EEG clips, three of which are non-seizure EEG clips. The non-seizure EEG clips exhibit significant differences in temporal features, motivating the opportunity to use them to increase class specificity.}
  \label{fig:dataset_overview}
\end{figure}

\subsection{Scaling training data with workflow notes} \label{subsec:scaling_training}

\paragraph{Seizure onset detection task.} Following previous studies \cite{saab2020weak,tang2022self}, our task of interest is to classify the existence of a seizure onset in a $60$ second EEG clip. Each EEG contains 19 electrodes that sample voltage readings at $200$ Hz, therefore the input to the model is a $60$-sec EEG clip $x \in \R^{12{,}000 \times 19}$ and the output is a binary label $y \in \{0,1\}$ indicating the existence of a seizure onset in that clip. To evaluate and compare the performance of deep learning models on the task of seizure-onset detection, we curated a gold-standard evaluation set of $626$ EEG hours ($37{,}588$ 60-sec EEG clips) labeled by two fellowship-trained EEG readers (details in Section \ref{sec:dataset}).

\paragraph{Workflow notes.} 
%A fundamental trend in supervised deep learning is that the ability of a model to generalize to unseen examples improves as we increase the scale of labeled training data \cite{}. 
Since acquiring gold-standard labels for all $68{,}920$ hours of EEG (or $4{,}135{,}225$ clips) would be extremely costly, we used a cost-effective technique that leverages workflow notes proposed by Saab et al. \cite{saab2020weak}. As visualized in Figure \ref{fig:figure1}, a standard clinical workflow for EEG analysis starts with (1) EEG data collection, after which, (2) a mixed group of (mostly) technicians, fellows, and students are tasked with annotating any salient events that may be useful for the final stage, where (3) fellowship trained clinical neurophysiologists give a final diagnosis in a written report. Importantly, while the annotators in the second stage are less skilled and may therefore make mistakes, their annotations are well suited for ML supervision due to their fine-grained structure and temporal specificity. In particular, the annotations consist of repetitive standard descriptions of events such as seizures, spikes, and movements, and are produced with precise timestamps of when each event occured. Since the annotations only contain the start times of events, we only label the clip in which that event began, and we do not label subsequent clips unless another (or different) event occurs.
The workflow notes are readily available for all EEGs in both our adult and pediatric hospitals, allowing us to scale our training data to unprecedented levels.

Each EEG recording is accompanied by a table of workflow notes, where each row is a logged note containing the text describing an event along with a timestamp representing the start of the event. 
We found $26$ relevant attributes from our manual analysis, and wrote simple regular expressions to extract the unique attributes from the workflow notes (e.g., considering synonyms and case-insensitivity; details in Section \ref{sec:dataset}). 
Figure \ref{fig:dataset_overview} displays a histogram of the $18$ most frequent attributes, where for example we have seizure onset annotations for $26{,}498$ EEG clips, spike annotations for $8{,}942$ EEG clips, and movement artifact annotations for $16{,}806$ EEG clips.

\paragraph{Impact of scaling labeled training data with workflow notes.}
We hypothesize that even though workflow notes may contain errors and our regular expressions may extract noisy labels, leveraging workflow notes to scale the training data results in better performing models compared to training models using a much smaller subset of gold-standard labels. 
To test our hypothesis, we randomly split our gold-standard labeled dataset into train ($50$\%), validation ($10$\%), and test ($40$\%) sets, stratified by patients (i.e., there are no overlapping patients among the three splits). We then trained two classification models, where the first model was trained using the gold-labeled train set (containing $16{,}058$ EEG clips, of which $408$ contained a seizure onset), and the second model was trained using the entire training set that was not gold-labeled, resulting in $4{,}097{,}637$ EEG clips, of which $25{,}254$ contained seizure onset labels extracted from the workflow notes. Details on model architecture and training procedure can be found in Section \ref{sec:methods}. To evaluate seizure onset detection performance, we assessed the Area Under the Receiver Operating Characteristic curve (AUROC) on the held-out test set, and report the $95$\% confidence intervals.

Leveraging the workflow notes improved the model's performance, where the model trained on the smaller gold-labeled dataset achieved an AUROC of $73.3 \pm 3.2$, and the model trained on the much larger workflow-labeled dataset achieved an AUROC of $85.6 \pm 0.9$.

\subsection{Revealing underperforming subgroups}  \label{subsec:revealing_subgroups}

To evaluate whether our models performed less well in certain patient subgroups, we performed a subgroup analysis where we evaluated the change in model performance across multiple clinically-relevant subgroups. 
We carried out the subgroup analysis by using a collection of patient metadata, gold-labeled seizure types, and the attributes from the workflow notes. 

For patient subgroups, we recorded whether the patient was from the adult or pediatric hospital, and whether a patient's EEG recordings were collected in the ICU. For seizure subtypes, we analyzed performance differences among the focal spike-and-wave, evolving rhytmic slowing, and generalized spike-and-wave types (more details in Section \ref{sec:dataset}).

\begin{figure}[htb]
  \begin{subfigure}{0.6\textwidth}
    \centering
    % Table goes here
    \begin{tabular}{|c|c|c|}
    \hline
    \multicolumn{1}{|l|}{}                                                        & Subgroup                                           & \multicolumn{1}{l|}{AUROC}         \\ \cline{2-3} 
    \multicolumn{1}{|l|}{\multirow{-2}{*}{}}                                      & Overall                                            & 85.6 ± 0.9                         \\ \hline
                                                                                  & Adults                                             & 89.4 ± 1.1                         \\ \cline{2-3} 
                                                                                  & Adults outside ICU                                 & 89.4 ± 1.7                         \\ \cline{2-3} 
                                                                                  & Adults from ICU                                    & 88.5 ± 1.2                         \\ \cline{2-3} 
    \multirow{-4}{*}{\begin{tabular}[c]{@{}c@{}}patient\\ subgroups\end{tabular}} & \cellcolor[HTML]{ECF4FF}Pediatrics                 & \cellcolor[HTML]{ECF4FF}82.9 ± 1.5 \\ \hline
                                                                                  & \cellcolor[HTML]{ECF4FF}Focal spike-and-wave       & \cellcolor[HTML]{ECF4FF}84.3 ± 2.6 \\ \cline{2-3} 
                                                                                  & Evolving rhythmic slowing                          & 89.8 ± 3.3                         \\ \cline{2-3} 
    \multirow{-3}{*}{\begin{tabular}[c]{@{}c@{}}seizure\\ subgroup\end{tabular}}  & \cellcolor[HTML]{ECF4FF}Generalized spike-and-wave & \cellcolor[HTML]{ECF4FF}85.5 ± 4.0 \\ \hline
    \end{tabular}
    \caption{}
    \label{subfig:baseline_gold_subgroups}
  \end{subfigure}%
  \begin{subfigure}{0.3\textwidth}
    \centering
    % Bar plot goes here
    \includegraphics[scale=0.3]{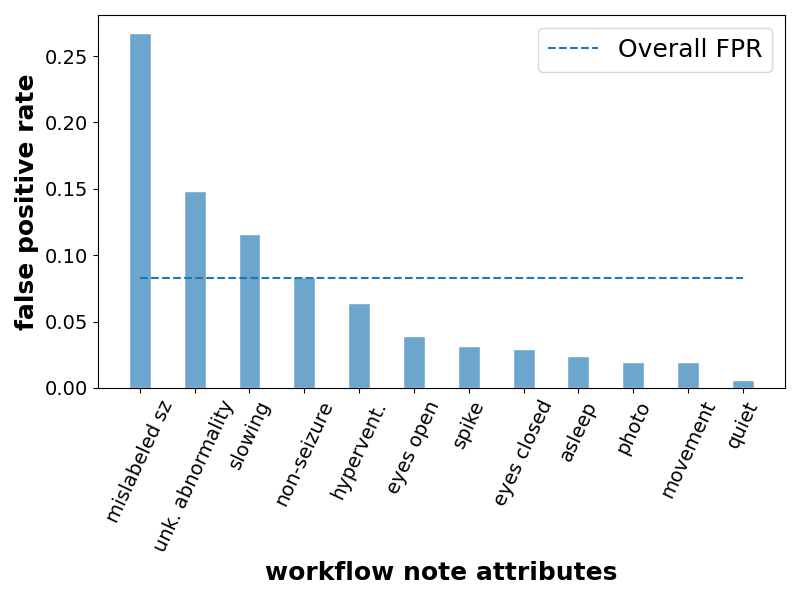}
    \caption{}
    \label{subfig:baseline_attribute_fprs}
  \end{subfigure}
  \caption{\textbf{Subgroup analysis:} (a): Model classification performance (AUROC with $95\%$ confidence intervals) for both patient and seizure subgroups. (b): False positive rate among workflow attributes.}
  \label{fig:baseline_analysis}
\end{figure}

From our subgroup analysis on patient and seizure types in Figure \ref{subfig:baseline_gold_subgroups}, we find that our model performed better for patients from the adult hospital with a $6.5$ AUROC point difference compared to patients from the pediatric hospital. There were also differences in the performance of the model for various seizure types, with a $5.5$ AUROC point difference between focal spike-and-wave and evolving rhythmic slowing seizures. From our subgroup analysis on workflow attributes in Figure \ref{subfig:baseline_attribute_fprs}, our model had significantly higher false positive rates (FPR) with respect to seizure onset detection for EEG clips with non-seizure abnormalities (FPR of $0.27$) compared to overall EEG clips (FPR of $0.08$). Details on metrics can be found in Section \ref{sec:metrics}.

\subsection{Improving subgroup robustness with class specificity} \label{subsec:improving_subgroups}

We hypothesize that our model underperforms on clinically-relevant subgroups as a result of the target task being underspecified. 
Since we train our model to only classify whether an EEG clip contains a seizure onset or not, 
all abnormal patterns and artifacts are grouped together with normal brain activity patterns (in the non-seizure class). As a result, unlike the training protocols of expert EEG readers, our model does not learn to differentiate among normal activity, abnormal seizure-like activity, and actual seizures, which we hypothesize causes the systematic errors displayed in Figure \ref{fig:baseline_analysis}. 

To combat task underspecification, we propose to train a multilabel model, where instead of outputing a binary class (seizure or non-seizure), the model identifies multiple attributes from an EEG clip, such as spikes, slowing, and movement. Importantly, since the workflow notes provide these attributes, we are able to train our multilabel model at no additional annotation cost, and training the model to recognize the additional attributes provides class specificity that we hypothesize can improve model performance. 
To test our hypothesis, we compared the overall and subgroup performances of a model supervised with binary seizure/non-seizure labels, which we will refer to as the binary model, to the same model trained on the same data but trained to classify the $26$ attributes (including seizure onset) extracted from workflow notes, which we will refer to as the multilabel model. While the multilabel model outputs probabilities for all $26$ attributes, we only consider the probability of seizure onset for evaluation, and calculate the AUROC with respect to the gold-labeled test set for each subgroup. 

\begin{table}[h]
\begin{center}
\begin{tabular}{cc|c|c|c|}
\cline{3-5}
                                                                                                    &                                                    & binary model                       & \textbf{multilabel model}                   & p-value \\ \hline
\multicolumn{2}{|c|}{Overall}                                                                                                                            & 85.6 ± 0.9                         & \textbf{91.5 ± 0.9}                         & 1.9e-24 \\ \hline
\multicolumn{1}{|c|}{}                                                                              & Adults                                             & 89.4 ± 1.1                         & \textbf{92.7 ± 1.1}                         & 7.9e-07 \\ \cline{2-5} 
\multicolumn{1}{|c|}{}                                                                              & Adults outside ICU                                 & 89.4 ± 1.7                         & \textbf{91.7 ± 1.7}                         & 0.036   \\ \cline{2-5} 
\multicolumn{1}{|c|}{}                                                                              & Adults from ICU                                    & 88.5 ± 1.2                         & \textbf{94.0 ± 1.2}                         & 1.1e-09 \\ \cline{2-5} 
\multicolumn{1}{|c|}{\multirow{-4}{*}{\begin{tabular}[c]{@{}c@{}}patient\\ subgroups\end{tabular}}} & \cellcolor[HTML]{ECF4FF}Pediatrics                 & \cellcolor[HTML]{ECF4FF}82.9 ± 1.5 & \cellcolor[HTML]{ECF4FF}\textbf{91.2 ± 1.3} & 5.1e-20 \\ \hline
\multicolumn{1}{|c|}{}                                                                              & \cellcolor[HTML]{ECF4FF}Focal spike-and-wave       & \cellcolor[HTML]{ECF4FF}84.3 ± 2.6 & \cellcolor[HTML]{ECF4FF}\textbf{92.0 ± 1.7} & 9.0e-06 \\ \cline{2-5} 
\multicolumn{1}{|c|}{}                                                                              & Evolving rhythmic slowing                          & 89.8 ± 3.3                         & 93.2 ± 3.0                                  & 0.10    \\ \cline{2-5} 
\multicolumn{1}{|c|}{\multirow{-3}{*}{\begin{tabular}[c]{@{}c@{}}seizure\\ subgroup\end{tabular}}}  & \cellcolor[HTML]{ECF4FF}Generalized spike-and-wave & \cellcolor[HTML]{ECF4FF}85.5 ± 4.0 & \cellcolor[HTML]{ECF4FF}90.0 ± 4.3          & 0.11    \\ \hline
\end{tabular}
\end{center}
\caption{\textbf{Improving subgroup robustness with class specificity:} Increasing class specificity improves overall model performance along with robustness to hidden subgroups. We stratified our evaluation set by patient and seizure subgroups, where the patient subgroups included patients from the adult hospital, pediatric hospital, or adults within or outside the ICU. We report the average AUROC along with 95\% confidence intervals. Rows highlighted in blue indicate subgroups that the binary model underperformed on. We estimated the p-value using the DeLong test, which evaluates how statistically significant the improvements of the multilabel model are compared to the binary model.}
\label{table:multilabel_subgroup}
\end{table}

As shown in Table \ref{table:multilabel_subgroup}, the multilabel model has significant improvements in both overall performance and subgroup performance (except for 2 of the seizure subgroups). The overall performance improved by $5.9$ AUROC points, while the performance on patients from the pediatric hospital improved by $8.3$ points, and $7.7$ AUROC points for focal spike-and-wave seizure types. Importantly, the improvements in performance significantly minimized the gaps in performance among subgroups. Additionally, we compared the FPRs for each attribute (shown in Supplementary Figure \ref{fig:comparison_attribute_fprs}) and found that the overall FPR decreased from $0.08$ to $0.02$. The top 3 attributes with the highest FPR, which correspond to abnormal attributes (mislabeled seizure, unknown abnormality, and slowing), all decreased significantly (e.g., FPR for EEG clips with unknown abnormal patterns decreased from $0.15$ to $0.08$). We further compared the 2D projected embeddings of the binary and multilabel models in Supplementary Figure \ref{fig:embedding_comparison}, which shows that the embeddings of the multlabel model of abnormal EEG clips cluster more tightly than the embeddings of the binary model, reaffirming that the multilabel model can better differentiate EEG abnormalities.

We also investigated the impact of training a multilabel model on different subsets of the workflow attributes on subgroup robustness. We choose two additional subsets of classes: classifying seizures along with two abnormalities highly relevant to seizures (spikes and slowing), and classifying seizures along with only abnormal attributes (i.e., we remove the following attributes: drowsy, jerk, tap, respiration, eyes open/closed, asleep, ekg, arousal). As shown in Supplementary Table \ref{table:multiple_multiclass}, we first found that all multilabel models improved overall seizure detection performance over the binary model. Interestingly, training a multilabel model for detecting seizure onset along with only abnormal attributes performed similarly to the multilabel model trained on all attributes, indicating that increasing class specificity with the abnormal attributes is the most important.

\subsection{Measuring clinical utility} \label{subsec:clinical}

%Studying metrics such as AUROC and FPR over a collection of EEG clips is useful in measuring the classification ability of a model. However, how does the classification ability of a model translate to measuring clinical utility? To measure the clinical utility of a seizure detection model, our evaluation needs to mimic realistic deployment settings and track metrics that directly impact a clinicians experience when utilizing a model's predictions.

A major barrier for technicians and neurophysiologists who have access to commercial seizure detection models is the high number of false alarms\cite{reus2023automated,pavel2020machine}, which results in alarm fatigue and in clinicians not utilizing model predictions. Therefore, a good metric to assess clinical utility is the average number of false positives after scanning $24$ hours of EEG (FPs/24hr). In particular, we look at two parameters that directly impact the number of false positives:

\begin{itemize}
    \item Recall (or sensitivity): Specifying the desired recall implicitly determines the threshold used to binarize the seizure probabilities. While having a higher desired recall is advantageous (since we miss fewer seizures), it is in direct tension with false positives, where number of false positives increase as we increase recall. In some settings, such as counting the precise number of occurring seizures, it may be critical to have a high recall. While in other settings, where the model is used as an assistant to prioritize which parts of the EEG to read first, having a high recall is not as critical. For these reasons, we look at the FPs/24hr for a recall of $0.5$, $0.8$, and $0.9$.
    \item Delay tolerance ($\Delta_t$): we define the delay tolerance to be the maximum amount of time allowed between the actual seizure onset and the predicted seizure onset. In other words, if the time between actual and predicted seizure onset ($T$) is greater than $\Delta_t$, we count the predicted seizure as a false positive; however, if $T<\Delta_t$ then we count the predicted seizure as a true positive. The delay tolerance is an important parameter because not only does it impact how we determine the difference between a true or false positive, but it is also implicitly related to seizure detection latency --- the speed in which our model flags seizures. Seizure detection latency may be critical in some settings, for example if we would like to precisely localize the seizure onset region for patients in the epilepsy monitoring unit, it is critical we accurately analyze the EEG near the true onset zone before spreading occurs. In other settings, such as counting number of seizures, seizure detection latency is not a critical parameter. For these reasons, we look at the FPs/24hr for a delay tolerance of $1$ minute and $5$ minutes.
\end{itemize}

\begin{figure}[h]
  \centering
  \includegraphics[width=0.95\textwidth]{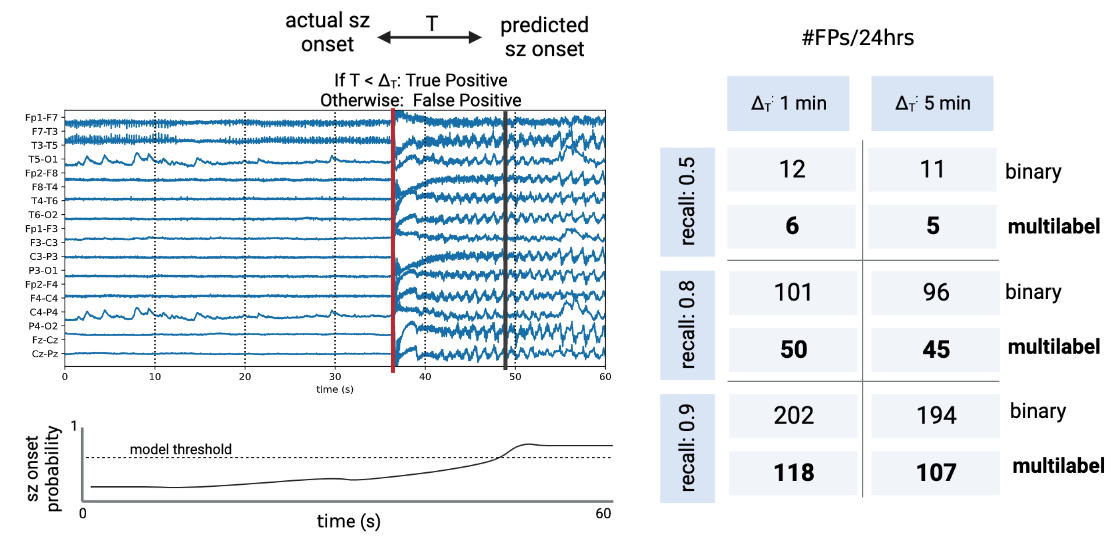}
  \caption{\textbf{Clinical utility metric:} On the left, is an EEG in which the red line indicates the actual seizure (sz) onset, and the black line indicates the predicted seizure onset by the model. The time elapsed between the actual and predicted onset is T, which is used to decide whether the predicted seizure onset is a true positive or false positive (depending on the delay tolerance for the clinical setting). The bottom left plot displays the model seizure onset probabilities across time, where the dashed line indicates the model threshold that is determined by the desired recall value. On the right, we compare the average number of false positives that occur every 24 hours of EEG in six different settings: a threshold such that we achieve a recall of $0.5$, $0.8$, or $0.9$, with either of two values of $\Delta_T$, which is a proxy to seizure detection latency (the maximum time between the ground truth and predicted seizure onset we tolerate).}
  \label{fig:clinical}
\end{figure}

In Figure \ref{fig:clinical}, we compared the FPs/24hr for 6 different settings while varying recall and delay tolerance, and observed that the multilabel model improved our clinical utility metric by a factor of roughly $2\times$ across all settings.

\section{Discussion}

In this work, we presented a strategy to improve the trustworthiness of seizure detection models by scaling training data and class specificity in a cost-effective manner.
Unlike existing techniques that require fellowship-trained neurophysiologists to annotate thousands of EEGs \cite{jing2023development}, we instead leveraged annotations that provide class specificity and are generated in existing clinical workflows \cite{saab2020weak}, allowing us to scale training data to an unprecedented level of $68{,}920$ EEG hours at no additional annotation cost. 
In addition to bypassing expert labeling of the training set, workflow notes can also facilitate the ongoing training of healthcare models as additional data are accumulated over time, leading to significant cost savings in terms of upfront and maintenance expenses.

Aside from annotation costs, a major roadblock to successfully deploying healthcare AI is the limited understanding of their errors within hidden subgroups of patients, leading to a lack of trust \cite{reus2023automated,pavel2020machine,wiens2019no}. 
Currently, the gold-standard technique to conduct an in-depth error analysis requires a clinician to manually interpret multiple data samples that the model classified incorrectly and find patterns that combine errors into clinically-relevant subgroups. 
Instead, we proposed to utilize patient metadata, gold-labeled seizure types, and multiple attributes describing EEG events to stratify the evaluation set and analyze differences in model performance. Apart from the gold-labeled seizure types, we are able to extract the attribute labels from the workflow notes, allowing us to greatly improve performance of our model with no additional costs. 
From our subgroup error analysis, we found that binary seizure classification models may have large performance gaps among patient age groups (-$6.5$ AUROC points on pediatrics compared to adults), seizure types (-$5.5$ AUROC points on focal spike-and-wave versus evolving rhythmic slowing), and has significantly higher false positives (+$19$ FPR points) for non-seizure EEG clips with abnormal brain activity compared to non-seizure clips. 
Identifying underperforming subgroups is a critical step in our goal towards trustworthy seizure classification models. 

Our core hypothesis is that our binary classification model has high false positives on abnormal non-seizure EEG clips as a result of task underspecification. Since fellows are not only trained to differentiate seizure from non-seizure activity, but also to identify multiple artifacts and abnormalities to rule out seizure \cite{tatum2021handbook}, we reason that a model should similarly be trained. To reduce high false positive rates and systematic errors, we leveraged attribute labels extracted from workflow notes and trained a multilabel model that learns to classify $26$ EEG events such as seizures, spikes, slowing, and movement. We found that such a multilabel model significantly improves overall performance (+$5.9$ AUROC points), along with closing the performance gap among subgroups, and decreased the false positive rate on abnormal non-seizure clips by $8$ FPR points, compared to the binary classification model. 
We believe this general direction of increasing the specificity of the supervision task is a promising approach to improve model subgroup robustness. Other successful approaches within this direction include increasing spatial specificity for radiology\cite{saab2022reducing} (e.g., segmentation) and training a chest X-ray model with a comprehensive class ontology \cite{seah2021effect}.

In our investigation of seizure detection models, we also establish a metric of clinical utility. We report the average number of false positives per $24$ hours of EEG for different recall and latency settings. We found that across different clinical settings, increasing class specificity reduces the FPs/24hr by a factor of $2\times$, suggesting that our improvements in subgroup robustness may translate to improvements in clinical utility.

Our proposed supervision strategies for improving trustworthiness of seizure detection models have limitations. 
First, while workflow notes offer a great alternative to manual expert labeling, the resulting labels come from personnel that are instructed to bias their reading to not miss abnormalities since final diagnosis is reviewed by an interpreting physician, which results in false positive labels and sub-optimal supervision. 
Additionally, our regular expressions to extract labels from the workflow notes may not correctly identify some of the labels, or they may produce errors or not apply to other institutions. 
Second, while we consider many clinically-relevant subgroups, our analysis can be more comprehensive by including many other important groups such as patient demographics, more seizure types, and finer-grained abnormal events. 
Third, we do not investigate other important robustness settings that include common distribution shifts, such as different EEG devices and patients from multiple hospitals. 
Other settings for improving trust may also include proper model calibration, calibration scores, and out-of-distribution detection. 
We believe it is critical to investigate robustness on a comprehensive list of settings before claiming a model to be trustworthy for deployment.

Future work is needed for improving the robustness of seizure detection models. 
Further scaling training data to include diverse patients can be done by combining our hospital datasets with existing publicly available datasets such as the TUSZ corpus\citep{obeid2016temple, shah2018temple}.
In a similar spirit, we can utilize publicly available EEG-based models that classify seizures, sleep staging, and brain states \cite{tang2022spatiotemporal,bashivan2015learning}, to either label relevant attributes or enable transfer learning. 
Another exciting direction is self-supervised and generative AI, where models do not rely on labeled training data to learn useful data representation. For example, recent work has shown that pretraining to forecast EEG signals boosts performance on rare seizure types\cite{tang2022self}. 
We also envision models that generate text reports from EEG \cite{biswal2019eegtotext} may prove to have more robust representations due to learning finer-grained concepts.

In summary, our work provides evidence that scaling training data using labels from workflow notes and increasing class specificity are promising techniques to improve robustness of models to detect seizure onset. 
We believe that combating robustness challenges through in-depth error analyses, and assessing detection performance of models as well as clinical utility metrics, will be critical to continue improving upon the trustworthiness of AI tools for clinical deployment.

\section{Methods} \label{sec:methods}

\subsection{Dataset description} \label{sec:dataset}

Our dataset consists of all EEGs recorded in both the Stanford Hospital and Lucile Packard Children's Hospital from 2006 to 2017. In total, our dataset contains $68{,}920$ EEG hours from $12{,}297$ patients. Our dataset is diverse, where patients span all ages, come from different hospital locations (ICU, epilepsy monitoring unit, and ambulatory), and have different seizure types and etiologies. More details on the statistics of our diverse patients can be found in Figure 2 and Supplementary Figure 2 in Saab et al.\cite{saab2020weak}.
 
To prepare input data samples from long-form EEG recordings, we segment each recording into non-overlapping $60$ second clips (i.e., stride is $60$ seconds). 
In total, our dataset contained $4{,}125{,}225$ clips. 
%Since we are only provided with the start time of seizures from the workflow notes, we do not consider EEG clips between seizure onset times, as we do not know 
To ensure consistent information across patients, we only considered the $19$ electrodes from the standard 10-20 International EEG configuration, and exclude premature infants or patients with small heads that prevent the full deployment of the 19 electrodes. 
We further normalize each EEG clip across the temporal dimension using the global average and standard deviations for each channel. Such normalization of input samples is standard practice in deep learning and we find this improves training. 

\paragraph{Gold-standard annotations.} Two fellowship-trained EEG readers (M.T. and C.L.M.) interpreted a randomly selected subset of EEG recordings, annotating for seizure onset. This resulted in an evaluation set of $37{,}588$ 60-sec EEG clips (or $626$ EEG hours), of which $1{,}244$ clips contain seizures from $395$ patients. Patients in the the evaluation set are excluded from the training set. 
C.L.M. labeled or supervised the labeling of each EEG clip according to the seizure type as defined by EEG ictal patterns; specifically, whether a seizure was a focal spike-and-wave, evolving rhythmic slowing, generalized spike-and-wave, paroxysmal fast acivity, polyspike-and-wave (myocolonic), or electrographically silent, for a subset of $358$ patients from the gold-labeled EEGs. However, due to the low frequency of some seizure types, our evaluations only included focal spike-and-wave, evolving rhytmic slowing, and generalized spike-and-wave types (more details can be found in Supplementary Table \ref{table:seizure_types}).

\paragraph{Extracting labels from workflow notes.}
Each EEG recording is accompanied with a table of workflow notes with each row indicating an event description along with the event start time. 
However, the event descriptions are free-form text, and while the workflow annotators use repetitive and standard descriptions, there may be slight deviations. 
M.T. and C.L.M. analyzed the most common $1{,}000$ event descriptions and by consensus
determined a set of unique attributes that (1) are visibly detectable on EEGs, and (2) are typically used when searching for seizures (e.g., common artifacts that must be ruled out such as movement, or other abnormalities such as spikes and slowing). From this manual analysis, we found $26$ class attributes of interest. 
To classify whether one of the $26$ class attributes is mentioned in the event description, we produce simple regular expressions that take into account synonyms or acronyms the annotators may use. 
For example, an annotator may write ``seizure'', ``sz'', ``spasm'', or ``absence''; another example is the description of an unknown abnormality, which may simply be indicated by ``x'', or ``xx''. 
We provide a full list of the regular expressions used in Supplementary Table \ref{table:regexs}.

\subsection{Model architecture and training}

There have been many deep learning model architectures proposed for seizure classification, such as convolutional models (CNNs) \cite{o2020neonatal, raghu2020eeg,ievsmantas2020convolutional,saab2020weak}, recurrent neural networks (RNNs) \cite{vidyaratne2016deep,golmohammadi2017gated,aliyu2019epilepsy}, graphical neural networks (GNNs) \cite{tang2022self,varatharajah2017eeg,vo2022composing}, and more \cite{ahmedt2020neural,tang2022spatiotemporal,rasheed2020machine,siddiqui2020review,asif2020seizurenet}. 
In our work, we study the impact of training data scale and the specificity of the supervision task on seizure classification performance, and not model architecture.
However, due to inherent advantages of some architectures, such as simplicity and computational efficiency, we chose S4, a recently proposed convolutional-based model motivated by principles in signal processing\cite{gu2021efficiently}. 

\paragraph{Deep state space sequence model (S4).} 
The global architecture of S4 follows a similar deep learning architecture as the transformer encoder, in which each layer is composed of multiple filters, where each filter is a sequence-to-sequence mapping (mixing across time), followed by a non-linear activation function, followed by a linear layer (mixing across filters), and finally a residual connection. 
The major deviation from the transformer encoder is the sequence-to-sequence filter, which as opposed to an attention mechanism, is a one-dimensional convolutional filter parametrized by linear state-space models (SSMs).
An SSM is a fundamental model to represent signals and is ubiqutious across a range of signal processing and control applications \cite{kalman1960new,hamilton1994state}. A discrete SSM, which maps observed inputs $u_k$ to hidden states $x_k$, before projecting back to observed outputs $y_k$, has the following recurrent form: 
\begin{align}
    x_{k+1} &= \zA x_k + \zB u_k  \label{ssm_recurrent_1}\\
    y_k &= \zC x_k + \zD u_k  \label{ssm_recurrent_2}
\end{align}

Where $\zA \in \R^{d \times d}$, $\zB \in \R^{d \times 1} $, $\zC \in \R^{1 \times d}$, and $\zD \in \R$ are learnable SSM parameters, and $d$ is the dimension of the hidden state $x$.   
Importantly, we can also compute the SSM as a 1-D convolution, which unlike recurrent models, enables parallelizable inference and training. To see how, if we assume the initial state $x_0=0$, and follow equations \ref{ssm_recurrent_1} and \ref{ssm_recurrent_2}, we arrive at the following by induction:
\begin{equation}
    y_k = \sum_{j = 0}^{k - 1} \zC \zA^{k - 1 - j} \zB u_j 
\end{equation}

We can thus compute the output $y_k$ as a 1-D convolution with the following filter:
\begin{align}
    \zF &= (\zC\zB, \zC\zA \zB, \zC\zA^2 \zB, \ldots, \zC\zA^{\ell - 1} \zB)  \\
    y_k &= (\zF * \zu)_k
\end{align}

Following prior work on sequence model classification\citep{gu2021efficiently}, we simply use the output squences from the last layer to project from the number of filters to the number of classes (e.g., $2$ classes for the binary model and $26$ classes for the multilabel model), and perform mean pooling over the temporal dimension before a softmax to output class logits. 
% from seuqence to classification
% going from multilabel predictions to binary prediction 

There are many advantages of using deep SSMs for long sequence modeling described in recent work\cite{gu2021efficiently,zhang2023effectively,gu2021combining}. We highlight the following advantages for EEG modeling: since our EEG clips are of length $12{,}000$, RNNs are slow to train, while CNNs fail to capture long-range dependencies due to limited filter lengths; on the other hand, SSMs are computationally efficient to train (due to their convolutional view), but are also able to capture long-range dependencies with structured initialization of the $\zA$ matrix. Moreover, we do not need to learn graph structures among the EEG electrodes, which adds an additional layer of complexity in recent state-of-the-art EEG classification models \cite{tang2022spatiotemporal}. 
Nevertheless, to validate that S4 is a well suited model architecture for seizure classification, we compared its performance to other architectures on the public TUSZ benchmark in Supplementary Table \ref{fig:model_comparisons}, and found that S4 is competitive with state-of-the-art models while being more computationally efficient. 

\paragraph{Training details.}
We trained all models with the cross-entropy loss using the Adam optimizer in Pytorch \cite{kingma2014adam}, with randomly initialized weights. The learning rate was initially set at $0.004$ and followed a cosine scheduler \cite{loshchilov2016sgdr}. We used a weight decay of $0.1$ and a dropout probability of $0.1$.
Since the training set is very large ($\sim4$ million samples) and highly unbalanced with just $0.6\%$ of clips having seizure onset, we used a weighted random sampler with a $25$-to-$1$ bias for positively labeled clips. For more frequent checkpointing, we randomly sampled a maximum of $150{,}000$ clips for each epoch (with replacement), and trained for $200$ epochs, while checkpointing on the validation set AUROC. 
The S4 model architectures had a parameter count of $366$k for the binary classification model, and $379$k for the multilabel model (due to larger output dimension). The model architecture contained $128$ filters per layer for $4$ layers with a hidden state dimension $d$ of $64$, and the gaussian error linear unit for the non-linear activations.
We performed a grid search for the initial learning rate, weight decay, and dropout values using our validation set. We used default values for the other hyperparameters, including model architecture.

\subsection{Performance metrics} \label{sec:metrics}
The two main classification metrics used to evaluate seizure classification performance are the the Area Under the Receiver Operating Characteristic curve (AUROC) and the false-positive rate (FPR). 
We chose the classification threshold such that the class balance of the model predictions matches the ground truth class balance. 
The ROC curve displays the tradeoff between the True Positive Rate (TPR) and FPR for different classification thresholds. Therefore, the AUROC summarizes the ROC curve in a single scalar value regardless of the specific classification threshold chosen. The FPR and TPR are defined as follows:

\begin{align}
    FPR &= \frac{FP}{FP+TN}  \\
    TPR &= \frac{TP}{TP+FN}
\end{align}

where true-positives (TP) are correct seizure classifications, true-negates (TN) are correct non-seizure classifications, false-positives (FP) are incorrect seizure classifications, and false-negatives (FN) are incorrect non-seizure classifications.
To calculare $95\%$ confidence intervals and p-values when comparing the AUROC of two models, we used the DeLong test \cite{delong1988comparing}.

\section*{Data Availability}
The Stanford clinical datasets used in this study are subject to restrictions regarding the availability of Protected Health Information. They were accessed with approval from the Institutional Review Board solely for the purpose of this specific study and are not accessible to the public.  

\section*{Code Availability}
The code used to generate the main results in this manuscript can be found in the following github repository: \url{https://github.com/khaledsaab/eeg_robustness}.

\newpage
\bibliography{refs}
\section*{Acknowledgements}

We gratefully acknowledge the support of Neuroscience:Translate grant from Wu Tsai Neurosciences Institute; NIH under No. U54EB020405 (Mobilize), NSF under Nos. CCF1763315 (Beyond Sparsity), CCF1563078 (Volume to Velocity), and 1937301 (RTML); US DEVCOM ARL under No. W911NF-21-2-0251 (Interactive Human-AI Teaming); ONR under No. N000141712266 (Unifying Weak Supervision); ONR N00014-20-1-2480: Understanding and Applying Non-Euclidean Geometry in Machine Learning; N000142012275 (NEPTUNE); NXP, Xilinx, LETI-CEA, Intel, IBM, Microsoft, NEC, Toshiba, TSMC, ARM, Hitachi, BASF, Accenture, Ericsson, Qualcomm, Analog Devices, Google Cloud, Salesforce, Total, the HAI-GCP Cloud Credits for Research program,  the Stanford Data Science Initiative (SDSI), Stanford Interdisciplinary Graduate Fellowship, and members of the Stanford DAWN project: Facebook, Google, and VMWare. The U.S. Government is authorized to reproduce and distribute reprints for Governmental purposes notwithstanding any copyright notation thereon. Any opinions, findings, and conclusions or recommendations expressed in this material are those of the authors and do not necessarily reflect the views, policies, or endorsements, either expressed or implied, of NIH, ONR, or the U.S. Government. We are also grateful for the helpful feedback and discussions with Jared Dunnmon, Nandita Bhaskhar, Krish Maniar, Yixing Jiang, and Pranav Gurusankar.

\newpage
\section*{Supplement}

\begin{table}[H]
  \begin{center}
  \begin{tabular}{|l|l|}
  \hline
  \textbf{seizure type}        &  \textbf{seizure count}        \\ \hline
  focal spike-and-wave & $181$ \\ \hline
  evolving rhythmic slowing & $81$\\ \hline
  generalized spike-and-wave & $53$ \\ \hline
  polyspike-and-wave (myoclonic) & $22$ \\ \hline
  paroxysmal fast activity & $17$ \\ \hline
  fast spiking & $7$  \\ \hline
  sz without clear electrographic change & $2$ \\ \hline
  electrographically silent & $1$ \\ \hline
  \end{tabular}
  \end{center}
  \caption{\textbf{Seizure types:} Seizure counts for different seizure types as defined by EEG ictal patterns in a subset of our gold-labeled evaluation set.}
  \label{table:seizure_types}
  \end{table}

\begin{table}[H]
\small
\begin{center}
\begin{tabular}{|l|l||l|l|l|}
\hline
Subgroups                          & sz only                            & sz / spikes / slowing                       & sz / all abnormal attributes                & sz / all attributes                          \\ \hline
\cellcolor[HTML]{ECF4FF}Overall    & \cellcolor[HTML]{ECF4FF}85.6 ± 0.9 & \cellcolor[HTML]{ECF4FF}\textbf{88.4 ± 1.0} & \cellcolor[HTML]{ECF4FF}\textbf{91.4 ± 0.9} & \cellcolor[HTML]{ECF4FF}\textbf{91.5 ± 0.9}  \\ \hline
Adults                             & 89.4 ± 1.1                         & 89.6 ± 1.5                                  & \textbf{92.8 ± 1.0}                         & \textbf{92.7 ± 1.1}                          \\ \hline
Adults w/o ICU                     & 89.4 ± 1.7                         & 88.7 ± 2.5                                  & 91.1 ± 1.7                                  & \textbf{91.7 ± 1.7}                            \\ \hline
Adults w/ ICU                      & 88.5 ± 1.2                         & \textbf{91.2 ± 1.5}                         & \textbf{94.7 ± 1.1}                         & \textbf{94.0 ± 1.2}                          \\ \hline
\cellcolor[HTML]{ECF4FF}Pediatrics & \cellcolor[HTML]{ECF4FF}82.9 ± 1.5 & \cellcolor[HTML]{ECF4FF}\textbf{87.8 ± 1.5} & \cellcolor[HTML]{ECF4FF}\textbf{90.7 ± 1.4} & \cellcolor[HTML]{ECF4FF}\textbf{91.2 ± 1.3}  \\ \hline
\cellcolor[HTML]{ECF4FF}Focal      & \cellcolor[HTML]{ECF4FF}84.3 ± 2.6 & \cellcolor[HTML]{ECF4FF}\textbf{89.4 ± 2.4} & \cellcolor[HTML]{ECF4FF}\textbf{90.9 ± 2.0} & \cellcolor[HTML]{ECF4FF}\textbf{92.0 ± 1.7}  \\ \hline
Evolving slow                      & 89.8 ± 3.3                         & 91.4 ± 4.1                                  & \textbf{94.8 ± 2.3}                         & 93.2 ± 3.0                                     \\ \hline
Generalized                        & 85.5 ± 4.0                         & 73.7 ± 7.5                                  & 89.9 ±  3.5                                 & 90.0 ± 4.3                                     \\ \hline
\end{tabular}
\end{center}
\caption{\textbf{Subgroup robustness:} Increasing task specificity improves overall model performance along with robustness to hidden subgroups. We stratify our evaluation set by patient and seizure subgroups, where the patient subgroups include patients from the adult hospital, children hospital, or adults within or outside the ICU. We report the average AUROC for the multilabel seizure detection model along with 95\% confidence intervals. Bolded numbers indicate statistically signficant lifts over the binary classification model. ``sz'' stands for seizure.}
\label{table:multiple_multiclass}
\end{table}

%The p-value evaluates how statistically significant the AUROC's are between the first column (lowest task specificity) and the last column (highest task specificity) using the DeLong test.

\begin{figure}[H]
    \centering
    % Bar plot goes here
    \includegraphics[scale=0.5]{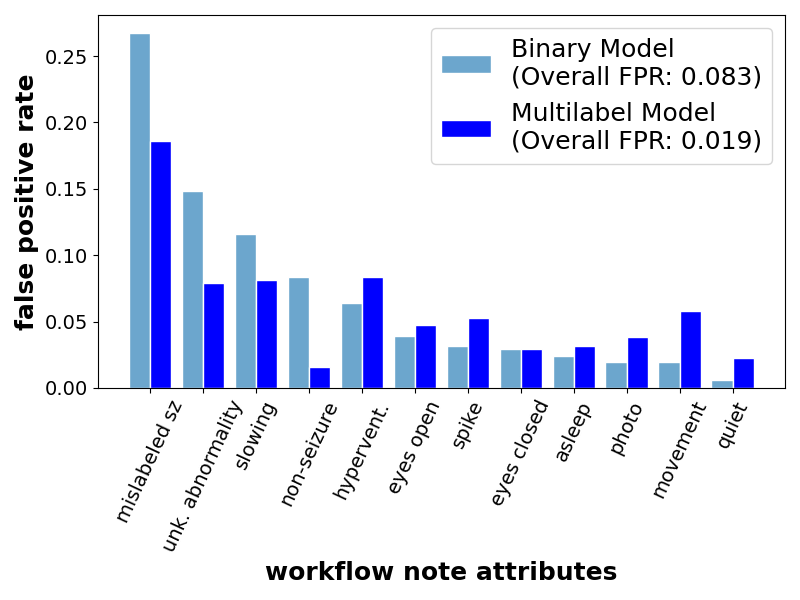}
  \caption{\textbf{FPR attribute analysis:} False positive rates with respect to seizure detection across subsets of our evaluation set stratified by the workflow attributes for the binary and multilabel model. The seizure detection threshold was chosen such that the class balance of the model predictions matched the ground truth class balance.}
  \label{fig:comparison_attribute_fprs}
\end{figure}

\begin{figure}[H]
\centering
\begin{subfigure}{0.4\textwidth}
\includegraphics[width=\linewidth]{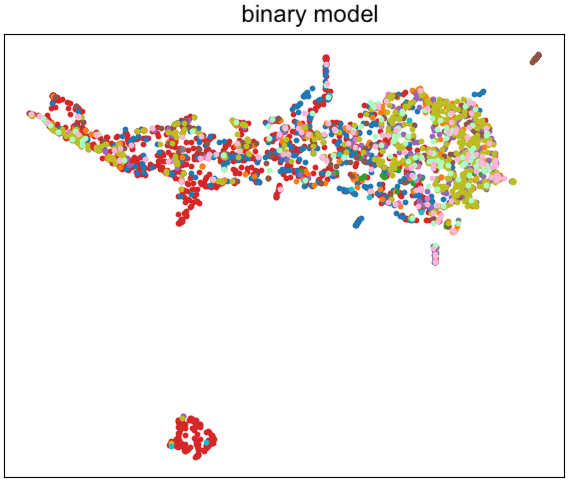}
\end{subfigure}
\hfill
\begin{subfigure}{0.55\textwidth}
\includegraphics[width=\linewidth]{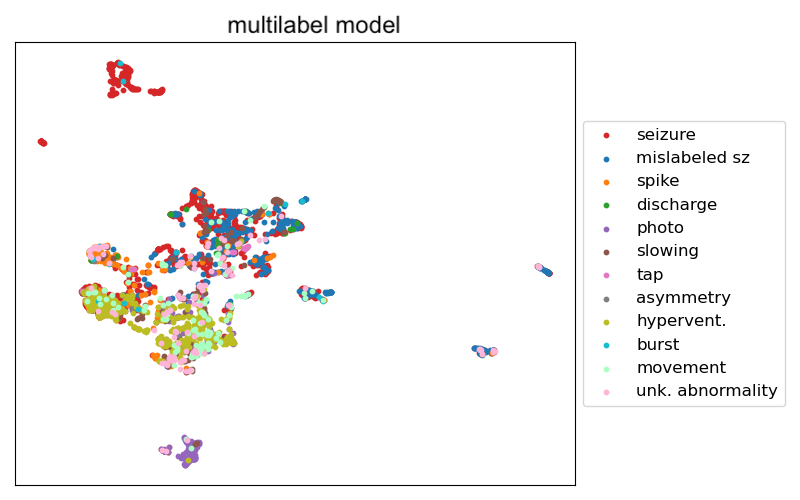}
\end{subfigure}
\caption{\textbf{Model embedding analysis:} UMap-projected embeddings show that the multilabel model embeddings cluster the abnormal attributes (mislabeled sz, spike, slowing, unknown abnormality) more tightly compared to the binary model embeddings, reaffirming that the multilabel model has learned to more effectively differentiate seizures from other EEG abnormalities.}
\label{fig:embedding_comparison}
\end{figure}

\begin{table}[H]
  \begin{center}
  \begin{tabular}{|l|l|}
  \hline
      \textbf{attribute} &  \textbf{regular expression}    \\ \hline
      seizure &  ``seizure $\mid$ sz $\mid$ absence $\mid$ spasm''    \\ \hline
      spike & ``spike''  \\ \hline
      slowing & ``slow''  \\ \hline
      photoelectric stimulation & ``photo''  \\ \hline
      stimulation & ``stim''  \\ \hline
      posterior dominant rhythm& ``pdr''  \\ \hline
      unknown abnormality & ``\^{}x*\$''  \\ \hline
      movement artifact & ``movement $\mid$ mvt''  \\ \hline
      EKG artifact & ``ekg''  \\ \hline
      discharge & ``discharge $\mid$ discharges''  \\ \hline
      tapping artifact & ``tap''  \\ \hline
      hyperventilation & ``hv''  \\ \hline
      jerking & ``jerk''  \\ \hline
      drowsy & ``drowsy''  \\ \hline
      asymmetry & ``asymmetry''  \\ \hline
      arousal & ``arousal''  \\ \hline
      respiration & ``respiration''  \\ \hline
      asleep & ``asleep $\mid$ sleep''  \\ \hline
      awake & ``awake''  \\ \hline
      burst & ``burst''  \\ \hline
      quiet & ``quiet''  \\ \hline
      suspicion in left hemisphere & ``\^{}L*\$''  \\ \hline
      suspicion in right hemisphere & ``\^{}R*\$''  \\ \hline
      eyes closed & ``eyes closed''  \\ \hline
      eyes opened & ``eyes opened''  \\ \hline
  \end{tabular}
  \end{center}
  \caption{\textbf{EEG attributes and regular expressions to extract EEG attributes from workflow notes:} For each regular expression, we turn off the case sensitivity flag.}
  \label{table:regexs}
  \end{table}

  \begin{table}[H]
    \begin{center}
    \begin{tabular}{|l|l|l|l|l|l||l|}
    \hline
    Model        & LSTM       & CNN-LSTM   & Dense-CNN  & DCRNN      & Graphs4mer & S4         \\ \hline
    TUH & 71.5 ± 1.6 & 68.2 ± 0.3 & 79.6 ± 1.4 & 80.4 ± 1.5 & 90.6 ± 1.2 & 87.7 ± 1.1 \\ \hline
    \end{tabular}
    \end{center}
    \caption{Architecture comparisons on TUH v1.5.2\citep{shah2018temple} test set (AUROC). Our chosen architecture (S4) is competitive with SoTA seizure detection methods. Performance of baseline models (first five columns) are taken from Tang et al., 2022 \citep{tang2022spatiotemporal}.}
    \label{fig:model_comparisons}
    \end{table}

\end{document}